\pdfoutput=1
\documentclass[11pt]{article}

\usepackage{EMNLP2023}

\usepackage{times}
\usepackage{latexsym}
\usepackage{gb4e}
\usepackage{amsmath}
\usepackage{amssymb}
\usepackage{bm}
\usepackage{paralist}
\usepackage{placeins}
\noautomath

\usepackage{setspace}
\usepackage{tcolorbox}
\usepackage{soul}
\setstcolor{red}
\usepackage{todonotes}
{\begin{tcolorbox}[colframe=red!75!black]}%
  {\end{tcolorbox}}

\usepackage[T1]{fontenc}
\usepackage[utf8]{inputenc}
\usepackage[english]{babel}
\usepackage[english=british]{csquotes}

\usepackage{microtype}

\usepackage{inconsolata}

\usepackage{enumitem}
\usepackage{booktabs}
\usepackage{multirow}
\usepackage{xspace}
\usepackage[normalem]{ulem}
\useunder{\uline}{\ul}{}

\newcommand{\xtime}{\textsc{x-time}}
\newcommand{\xcountry}{\textsc{x-country}}
\newcommand{\xlmen}{\textsc{xlm-enc}}
\newcommand{\mt}{\textsc{mt}}

\title{Multilingual estimation of political-party positioning:\\
From label aggregation to long-input Transformers}
\author{Dmitry Nikolaev \quad Tanise Ceron \quad Sebastian Pad{\'o} \\
Institute for Natural Language Processing, University of Stuttgart \\
    \texttt{dnikolaev@fastmail.com} \quad \texttt{\{tanise.ceron,pado\}@ims.uni-stuttgart.de}}
\begin{document}

\maketitle

\begin{abstract}
    % % One of the classical tasks in computational political science is the \textit{scaling analysis}
    % Scaling analysis is an approach in computational political science that assigns  
    % %of political actors: assignment of
    % to a person, a party, or a body of text 
    % (e.g., a parliamentary speech or an election manifesto) a numerical score according to some predefined 
    % political scale
    % such as Left--Right and Libertarian--Authoritarian. Until recently, such analyses were mostly 
    % done using bag-of-words methods, and experiments using more powerful word and sentence-embedding 
    % approaches only targeted relatively small datasets in a restricted number of countries. In this work, we propose the first large-scale analysis of party positioning across 41 countries and 27 languages.
    % Taking the full annotated dataset of the Comparative Manifesto Project, we experiment with
    % multilingual embeddings, machine translation, and different sentence-level label-aggregation
    % strategies in order to propose a method for a fully automated analysis of party positioning 
    % along major political scales. Despite the necessity to project the model
    % across languages and time periods, we achieve promising results in two different
    % analytical settings, showing that our method can be applied across unseen countries and election cycles. 
    % One of the classical tasks in computational political science is the \textit{scaling analysis}

Scaling analysis is a technique in computational political science
that assigns a political actor (e.g.\ politician or party) a score on a predefined 
% political 
scale based on a (typically long) body of
text (e.g.\ a parliamentary speech or an election manifesto). %Examples of political scales are Left--Right and Libertarian--Authoritarian. 
For example, political scientists have often used the left--right scale to systematically analyse 
political landscapes of different countries. 
NLP methods for automatic scaling analysis can find broad application provided they 
(i)~are able to deal with long texts and 
(ii)~work robustly across domains and languages.
In this work, we implement and compare two approaches to
% compare and contrast a range of models for 
automatic scaling analysis of political-party manifestos: 
% , ranging from 
label aggregation, a pipeline strategy relying on annotations of individual 
statements from the manifestos,
%corresponds to the traditional way of doing scaling analysis, 
and long-input-Transformer-based models, which compute
scaling values directly from raw text.
We carry out the analysis of the Comparative Manifestos Project 
dataset  across 41 countries and 27 languages and find that 
the task can be efficiently solved by state-of-the-art 
models, with label aggregation producing the best results. 
% , 
% long-input-Transformer models benefit from depending less on extensive annotations. 

\end{abstract}

\section{Introduction}
\label{sec:introduction}

A widely used tool in computational political science is the so-called
\enquote{scaling analysis}: a set of methods for representing political
platforms as numbers on a certain scale, such as left--right,
authoritarian--libertarian, or conservative--progressive
\citep{laver2003extracting,slapin2008scaling,diermeier2012language,
lauderdale2014scaling,barbera2015birds}. A~wide variety of scales have
been proposed in the literature, some based on political-theoretic
considerations \citep{jahn2011conceptualizing}, others more
data-driven
\citep{gabel2000putting,albright2010multidimensional,rheault2020word}.
% in order to analyse the full political spectrum.

One well-established scoring scheme of this kind is the Standard Right--Left Scale,
also known as the RILE score \citep{budge2013standard,volkens2013rile}. It was developed in the framework
of the Manifesto Research on Political Representation project (MARPOR), formerly 
known as the Comparative Manifestos Project (CMP),\footnote{\url{https://manifestoproject.wzb.eu/}}
which collects, annotates, and makes available a large collection of party platforms from
different countries. The RILE score is a deductive, first-principle-based method for describing
party positions geared towards the widest possible applicability across time
and countries \citep{budge2013standard}.
% , which
% claims as its advantages (i)~the ability to \enquote{draw on all the information available in the 
% data-set} thus cancelling out non-systematic error and giving an overview of the main political
% processes that the data represent; and (ii)~its being a substantively invariant measure whose
% numeric values always carry the same meaning 
% \citep[2]{budge2013standard}. 
For this very reason, it is rather conservative and inflexible and has
been repeatedly criticised \citep[see,
e.g.,][]{flentje2017mcss}. Despite this, it is widely used in
computational political science for model validation
\citep{rheault2020word}, as a dependent variable in regression
analyses \citep{greene2016diverse}, or as a basis for party-stance
analysis \citep{daubler2022scaling}.

A major practical drawback of the RILE score is the fact that it is computed based on the labels
manually assigned by MARPOR annotators to all statements in party manifestos (see Section~\ref{sec:cmp} for details). This procedure is expensive
and time consuming, which raises the question of whether we can adequately
approximate the RILE score using natural-language-processing methods, especially in a
multilingual setting. This will make it possible to efficiently perform analyses 
of political texts that have not yet been covered by the MARPOR project due to timing 
constraints (e.g.\ manifestos from upcoming elections), accidental gaps (e.g., Indonesia and the Philippines are not part of the dataset, and coverage of many countries, such as South Africa, is incomplete), or lack of resources (there are very few annotated 
manifestos from before 2000). 

This work is a first step in this direction. Our contributions are the following:

\begin{enumerate}
% \begin{compactenum}
    \item Previous works on computational analysis of party positioning
    targeted a limited number of texts from a single country 
    or several countries. We scale the analysis up to 41 countries and 27 languages, 
    including comparatively low-resource languages
    (such as Georgian and Armenian) that have not been tackled before.

    \item We contrast the label-aggregation approach, based on a
 statement-level classifier mimicking the work of a human annotator,
 with using long-input Transformer models predicting the scores
 directly from raw manifesto texts.
 
    \item In the label-aggregation setting, we further compare the performance of 
    multilingual-modelling-based and machine-translation-based approaches. While the former is more straightforward in the sense that 
    a single base model can be directly used without any preprocessing, MT systems are easier to train for 
    less-resource-rich languages, and only a single-language classifier is needed for predictions.

    \item We evaluate the generalisability of models regarding two dimensions: 
    local (moving to new countries) and temporal (moving from the past to the future). 
    These correspond to different real-life research scenarios. 
    We show that our methods deal reasonably with both cases. 
    
    % \item We evaluate our methodology on two political dimensions:
    % % \footnote{The terms \textit{dimension} and \textit{scale} are used interchangeably in this text.} 
    % Right--Left and Libertarian--Authoritarian \citep{bakker20133choice}.
    %another way of aggregating MARPOR labels, the Libertarian--Authoritarian scale \citep{bakker20133choice}.
% \end{compactenum}
\end{enumerate}

The paper is structured as follows:
In \S~\ref{sec:cmp}, we provide more information on the MARPOR annotations and on
how the RILE score is computed.
% Section~\ref{sec:problem-statement} presents 
The exact problem statement, different operationalization strategies, and
the experimental setup are presented in \S~\ref{sec:experimental-setup}, while the results
of the study are given in \S~\ref{sec:results}.
Additional discussion is provided in \S~\ref{sec:discussion}.
Section~\ref{sec:related-work} surveys related work. Section~\ref{sec:conclusion} 
concludes the paper and discusses directions for future research.
% and proposes future research.

\section{MARPOR categories and political scales}\label{sec:cmp}
% \section{\hspace{-.5em}MARPOR categories and political scales}\label{sec:cmp}

\paragraph{Categories} The annotations of the manifesto created in the framework of the Comparative 
Manifestos Project follow the project codebook \citep{marpor2020}. Each
statement of a given manifesto is annotated with a category
representing a specific policy domain (e.g. \textit{Military} or
\textit{Sustainability}). These categories can be identified via their
names and numbers (e.g., 103, \textit{Anti-Imperialism}).\footnote{See Appendix~\ref{app:marpor-categories} for the major categories with numbers.}

%, the last version of which was released in late 2020 \citep{marpor2020}. 

%The codebook describes the coverage (election programmes of parties that won one or two seats in respective national elections, depending on the region), the unit of analysis (a single party in a given election), metadata (pertaining to the party, election results, coder identity and status, etc.), content-analytical variables aggregated over labels for individual statements, and, most importantly, \textit{categories} for labelling these statements. 

%The domain fall into several high-level \textit{domains} (External Relations, Freedom and Democracy, Political System, Economy, Welfare and Quality of Life, Fabric of Society, and Social Groups), which serve to organise the coding system but do not form a basis for a proper classification.

%Some additional categories were introduced for Central and Eastern European categories,\footnote{E.g., 6011, \textit{The Karabakh Issue: Positive}, and 6072, \textit{Multiculturalism pro Roma: Positive}.} and over time some main categories obtained  more fine-grained \textit{sub-categories}, which can be aggregated into respective main categories  for a more coarse-grained analysis or when comparing with earlier versions of the annotation.

A~key feature of MARPOR categories is that they are not stance-neutral. Thus, category 201, 
\textit{Freedom and Human Rights}, or subtypes thereof, are assigned to \enquote{favourable mentions 
of importance of personal freedom and civil rights} \citep[12]{marpor2020}. Some
categories form binary oppositions (e.g.\ \textit{Constitutionalism: Positive} vs.\ 
\textit{Constitutionalism: Negative}), and some are purely one-sided (e.g.\ \textit{Freedom} and 
\textit{Democracy} have positive loadings and do not have negative counterparts). 
As a result, it is possible to derive inferences about
political stances of different parties from category counts alone. This provides a straightforward
operationalization of the political-science notion of \textit{issue salience}, which is commonly
used to analyse political positioning \citep{epstein2000salience}~-- the number of occurrences of a category correlates with how important it is for a party. 

In total, there are 143 different categories, with 56 major categories, 32 sub-categories of
the major categories, 54 additional categories, and the residual category 0.\footnote{In some  manifestos, special label \enquote{H} is attached to headings.
As we cannot reliably automatically identify headings in new texts, H's were converted to 0's throughout.}

\paragraph{Right--Left scale} A prominent way of analysing party positioning
%aggregating issue saliences 
is the Standard Right--Left Scale, a.k.a.\ RILE score
 \citep{budge2013standard,volkens2013rile}. Originally developed in
 the framework of the MARPOR project, it has been consistently used in
 its publications and remains a standard reference scale for party
 positioning, despite a number of proposals to improve or replace it,
 both using theory-based and data-driven approaches \citep[cf.\
 ][]{cochrane2015left,molder2016validity,flentje2017mcss}.
\begin{equation}
    \text{RILE} = \dfrac{R - L}{R + L + O}
    \label{eq:rile}
\end{equation}
Eq.~\ref{eq:rile} shows the formula for computing the RILE score based
on sets of categories defined by MARPOR as right-wing and left-wing,
respectively. The categories belonging to the right and left
sets are shown in Table~\ref{tab:rile-categories}.

\begin{table*}[t]
\centering
\small
\begin{tabular}{ll}
\cline{2-2}
Right emphasis &
  \begin{tabular}[c]{@{}l@{}}Military: Positive, Freedom, Human Rights, Constitutionalism: Positive, \\ 
  Political Authority, Free Enterprise, Economic Incentives, Protectionism: \\ Negative, Economic Orthodoxy, 
  Social Services Limitation, National Way \\ of Life: Positive, Traditional Morality: Positive, Law and Order, 
  Social Harmony\end{tabular} \\ \cline{2-2} 
Left emphasis &
  \begin{tabular}[c]{@{}l@{}}Decolonisation, Anti-imperialism, Military: Negative, Peace, Internationalism: \\ 
  Positive, Democracy, Regulate Capitalism, Market, Economic Planning, \\ Protectionism: Positive, Controlled Economy, Nationalisation, Social Services: \\ Expansion, Education: Expansion, Labour Groups: Positive\end{tabular} \\ \cline{2-2} 
\end{tabular}
\caption{The MARPOR categories used for calculating the RILE score.}\label{tab:rile-categories}
\end{table*}

\textit{R}(right) and \textit{L}(eft) are the percentages of statements 
labelled with categories from the two sets, and \textit{O} the percentage of
other statements. The range of RILE is $[-1, 1]$. Large absolute
values indicate extreme left and right programs, and values close to
zero correspond to centrist manifestos with a balanced 
program.\footnote{They could also arise from political programs
where most statements are associated with neither left nor right, but
such programs are rare in practice.}

\section{Methods}\label{sec:experimental-setup}

\subsection{Operationalization}\label{ssec:operationalisation}

\paragraph{Label aggregation}
As outlined above, we aim at automatically estimating positions of
political parties on the Left--Right scale. An approach that closely
mirrors the traditional MARPOR procedure would be to automatically
label the sentences in the manifestos with MARPOR categories and
aggregate them according to Eq.~\ref{eq:rile}. Unfortunately,
classifying the sentences is difficult, as we will show below. Reasons
include the large number of labels, their uneven distribution,
%within and across manifestos,
and the country-specific nature of manifestos.

However, the predicted categories arguably do not have to perfect --
 it may be sufficient for high-quality scaling analysis if the
 mistakes are uncorrelated so that, for example, the number of
 sentences mistakenly classified as left-leaning or neutral is close
 to the number of sentences mistakenly classified as right-leaning or
 neutral. One way to further raise the signal-to-noise ratio is to
 predict more high-level 
 % aggregate 
 labels. To compute the RILE score,
 we do not require specific categories, but only a three-way
 classification (R[ight], L[eft], O[ther]), which is much more
 tractable.
 % with current data and models. 
 This approach can be easily
 mapped into other dimensions as long as there is a list of categories
 from MARPOR belonging to both poles of the scale.

\paragraph{Direct prediction}
As an alternative, we can define a function % \(\mathcal{F}:
$\mathcal{T} \rightarrow [-1, 1]$ that directly maps a text to its
RILE score, and approximate it with a neural regression
model. Until recently, such an approach was infeasible due to the
restrictions on the input length in the state-of-the-art embedding
models: 512 or 1024 tokens depending on the model size, which is not
enough to analyse longer texts. However, a~new generation of
long-input Transformers (LITs) based on lightweight variants of the
self-attention mechanism increased the input limit to 4096 tokens or
more \citep{tay2021long}. This still does not give us a way to
compute a score for a whole text, but averages of RILE scores for
4095-token chunks of manifestos nearly perfectly correlate with
gold manifesto-level scores (Spearman's \(r > 0.99\)), which makes
by-chunk estimation a good proxy.

An additional motivation to pursue this avenue is provided by the fact
that it not only removes the need to classify the labels of individual statements 
but also saves researchers the effort to identify statements in the first place. 
This is a non-trivial problem as, according to the MARPOR codebook, any
sequence of words with a distinct meaning can be considered a
statement. E.g., a sentence \textit{All well-meaning citizens should
strive to maintain the world peace} can be construed as a single
example of the category \textit{Peace}, or \textit{all well-meaning
citizens} can be assigned its own label of \textit{Civic mindedness}.
%
%In
%this case, the annotations contain two separate statements with their
%respective labels.
% \footnote{This complicated coding scheme and the fact 
% that most texts were coded by a single person lead to somewhat noisy annotations 
% \citep{mikhaylov2008coder}.}
In line with previous work, our aggregation-based approach assumes
that statement boundaries are known, but in practice they will have to
be predicted together with the labels, or the coding scheme must be
simplified, e.g.\ by assigning a single \enquote{majority} label to
each sentence. By virtue of working with raw text spans, LITs do not
have to make such compromises.

% TODO: announce subsubsections

%\subsubsection{Label structure}\label{sssec:label-structure}

%Previous work has explored two approaches to overcoming this difficulty. One exploits the fact that categories are grouped into several large domains by adding hierarchical inductive bias to the training regime \citep{dayanik-etal-2022-improving} (this approach was shown to be especially effective for smaller datasets), and the other groups MARPOR categories following the principle of topical coherence, which sometimes goes against MARPOR's domain structure \citep{ceron-2022}.

\subsection{Problem settings}\label{sssec:problem-setting}

We consider two settings, corresponding to two different research
scenarios.  In the \textsc{leave-one-country-out} (\xcountry) setting,
we train the model on all data from $n-1$ countries (split into
training and development sets), and evaluate it one held-out country.
This corresponds to the situation when manifestos from a country not
yet covered by the MARPOR project, such as Indonesia, need to be
analysed.  This is repeated for all countries.

% Given the large amount of data, this setting limits the size of
% models we could employ, and the analysis for \xcountry\ is based on BERT-base-sized
% models. 

In the \textsc{old-vs.-new} (\xtime) setting, we train the model on all data from
before 2019 and evaluate it on the data from 2019--2021. This corresponds to
the situation when new data from an already covered country become
available.\footnote{Another application for this setting is the analysis of manifestos of smaller  parties that did not win any seats in previous elections and were not included in the dataset. The 
converse -- \textsc{new-vs.-old} -- would permit running a 
historical analysis of party positioning within a country. We have not addressed this scenario due to the scarcity of annotations from before 2000.}

\subsection{Dataset}\label{ssec:dataset}

We use the annotated subset of the latest release of the MARPOR dataset
\citep[version\ 2022a; ][]{lehmann2022marpordata} augmented with the separately curated
South American dataset \citep{lehmann2022marporsa}.\footnote{All data are available
on the project web page: \url{https://manifestoproject.wzb.eu/datasets}.}
We excluded manifestos annotated before the year 2000 to obtain a more uniform
training dataset. 
%(we do not address the task of analysing historical data in this work) 
Furthermore, to ensure comparability between two approaches to cross-lingual 
modelling~-- preprocessing using machine translation and using a multilingual encoder 
(see \S~\ref{ssec:mt-models} below)~-- we excluded languages for which no pretrained 
free NMT system was readily available. 
This leaves us with 1314 manifestos from 41 different countries  in 27 different 
languages.

In the \xcountry\ setting,
the rolling test set includes all of the data, while in the \xtime\ setting it is
much smaller (163,714 vs.\ 1,062,302 statements in the training set, i.e.\ around 13\%) and has a weaker geographical coverage: only 18 countries have manifestos from 2019 and later.

The data for LITs have the same train-test general splits, but sentences
in them were consecutively concatenated into text chunks of size no more than 4095 tokens (see Section \ref{ssec:operationalisation}),
% SP: this is the kind of info you put into a camera ready
%\footnote{More precisely, a token-length variable was associated with each chunk, which kept track of the number of tokens in it as produced by the tokeniser associated with a particular model. The exact length and number of chunks therefore differ slightly across models.}
with a RILE score computed for each chunk based on its gold MARPOR
labels. Chunks of size less than 1000 tokens were 
discarded.\footnote{Statistics of the datasets are shown in Tables~\ref{tab:dataset-stats} (by country) and \ref{tab:dataset-stats-languages} (by language)
in Appendix~\ref{app:dataset-stats}.}

\subsection{Models}\label{ssec:mt-models}

The MARPOR dataset is multilingual, which raises the challenge of language transfer.
The two current approaches in this case are using a multilingual encoder or 
machine translating all the data into the pivot language, usually English
% \citep{schafer-etal-2022-cross,litschko2022cross,srinivasan-choi-2022-tydip}.
\citep{litschko2022cross,srinivasan-choi-2022-tydip}.

\paragraph{Label aggregation} 
Here we experiment with both options.
For the \textsc{multilingual-encoder track} (\xlmen), we extract the representation
of the CLS token from XLM-RoBERTa base (in the \xcountry\ setting) and XLM-RoBERTa base and large
(in the \xtime\ setting).\footnote{The necessity to train 41 different models on the full
dataset in the \xcountry\ setting made it impractical to use the large model.}
Throughout, the classification head is a 2-layer MLP with the inner dimension of 1024
and tanh activation after the first layer.

In the \xcountry\ setting, the model was then repeatedly trained for two epochs using cross-entropy 
loss and the AdamW optimiser \citep{loshchilov2019adamw} with the learning rate of \(10^{-5}\).\footnote{The code for 
training and evaluating the models can be found at \url{https://github.com/macleginn/party-positioning-code}} In the \xtime\ setting, the general
setup is the same but the model was trained for five epochs with a checkpoint selected
based on the dev-set accuracy.

For the \textsc{machine-translation track} (\mt), all manifestos are translated into
English, for which the best MT systems and arguably the best pre-trained encoders are 
available. The current MT systems, however, are still rather noisy, especially for 
non-WEIRD \citep{henrich2010} languages, which offsets the benefits of a stronger base model.

We use the EasyNMT toolkit\footnote{\url{https://github.com/UKPLab/EasyNMT}} giving access to the Opus-MT models \citep{tiedemann2020opusmt}. A~cursory inspection of the translated sentences shows that the translation quality does vary across languages. However, even for manifestos whose source languages are difficult to translate (e.g.\ Georgian) the results produced by the classifier are still acceptable.

The translated sentences are encoded using pooled representations from
\texttt{all-mpnet-base-v2}, a version of MPNet \citep{song2020mpnet} fine-tuned 
following the SBERT methodology \citep{reimers-gurevych-2019-sentence} and available on 
HuggingFace.\footnote{\url{https://huggingface.co/sentence-transformers/all-mpnet-base-v2}
Preliminary experiments showed, in agreement with the results of \citet{ceron-2022}, 
that it consistently outperforms RoBERTa in monolingual settings.} The same 
classification head was then used as in the \xlmen\ approach, as well as
the same training parameters.

For each model, we aggregate the labels across manifesto sentences  
and compute its RILE score according to Eq.~\ref{eq:rile}.

\paragraph{Direct prediction} We experiment with two long-input encoder models: 
Longformer \citep{beltagy2020longformer} and BigBird \citep{zaheer2020big}.\footnote{Pretrained models
were downloaded from HuggingFace: \url{https://huggingface.co/allenai/longformer-base-4096} and
\url{https://huggingface.co/google/bigbird-roberta-base}.}
They are only available for English, and we apply them to the translated dataset. 
We use the embedding of the last layer's CLS token as input to a regression head. 
In the training step, each chunk receives a 
gold label computed from its sentences using Eq.~\ref{eq:rile}. 
% Given the nature of the labels, the LITs models are trained as a regression problem. 
The final RILE score of each manifesto is the average of regression values of its chunks. 
The regression head is similar to the classification head described above with the final softmax 
layer replaced with a single node with tanh activation mapping the output into the \([-1, 1]\) range. 
The systems are trained using MSE loss.  

\subsection{From regression to classification with LITs}

A possible concern about the direct computation of RILE scores, as we
frame the task for LITs, is that the models may fail to implicitly 
recreate the labelling-and-aggregation pipeline and instead learn 
spurious shortcuts by observing correlations between properties of texts 
and their RILE scores, which will then hurt test performance.

To address this concern, we carry out an additional experiment where
we make the models' task more comparable to what a human political
analyst would do. We train the LITs in a
binned-regression setting: the range of RILE scores is split into
five regions, corresponding to \textit{hard left} \([-1, -0.6)\),
\textit{centre left} \([-0.6, -0.2)\), \textit{centrist} \([-0.2,
0.2)\), \textit{centre right} \([0.2, 0.6)\), and \textit{hard right}
\([0.6, 1]\). The models are then trained to predict these classes
instead of real-valued RILE scores using cross-entropy loss.

\subsection{Evaluation metrics}\label{ssec:evaluation-metrics}

For the label-aggregation models, we first diagnose the performance of
the label classifiers using the weighted macro-averaged F1 score.

We then evaluate both the label-aggregation and the direct-prediction
models on the target task of predicting RILE score. We use Spearman's
correlation coefficient, which shows if our scores are monotonically
related to those computed from gold annotations using
Eq.~\ref{eq:rile}.  Additionally, we look at absolute values of errors
and their directionality.

We evaluate the performance of the LIT-based classifiers in the
binned-regression setting using accuracy and F1 score.

\section{Results}\label{sec:results}

The main results of the experiments are summarised in Tables~\ref{tab:results-categories} and \ref{tab:results-correlations}. Sections~\ref{ssec:results-categories} and \ref{ssec:results-scales} discuss the results while \S~\ref{ssec:results-errors} provides some detail about the strengths and weaknesses of the models.

\begin{table}[t]
\centering
\small
  \setlength{\tabcolsep}{4pt}

\begin{tabular}{@{}llrrrrrr@{}}
\toprule
 &     & \multicolumn{3}{c}{\textbf{\textsc{x-country}}} & \multicolumn{3}{c}{\textbf{\textsc{x-time}}} \\  
 &
   &
  \multicolumn{1}{c}{\textsc{xlm}} &
  \multicolumn{1}{c}{\textsc{mt}} &
  \multicolumn{1}{c}{\textsc{maj}} &
  \multicolumn{1}{c}{\textsc{xlm}} &
  \multicolumn{1}{c}{\textsc{mt}} &
  \multicolumn{1}{c}{\textsc{maj}} \\ \cmidrule(l){3-8}
 & Acc & 0.46        & 0.47        & 0.10       & 0.54       & 0.48       & 0.10       \\
\multirow{-2}{*}{CMP} &
  F1 & 0.44 & 0.44 & 0.02 & 0.55 & 0.48 & 0.02 \\ \cmidrule(l){2-8}
  % \cellcolor[HTML]{EFEFEF}0.44 &
  % \cellcolor[HTML]{EFEFEF}0.44 &
  % \cellcolor[HTML]{EFEFEF}0.02 &
  % \cellcolor[HTML]{EFEFEF}0.55 &
  % \cellcolor[HTML]{EFEFEF}0.48 &
  % \cellcolor[HTML]{EFEFEF}0.02 \\
 & Acc & 0.70         & 0.71        & 0.59       & 0.77       & 0.74       & 0.63      \\
\multirow{-2}{*}{RILE} &
  F1 & 0.70 & 0.70 & 0.44 & 0.77 & 0.74 & 0.49 \\ \bottomrule
  % \cellcolor[HTML]{EFEFEF}0.7 &
  % \cellcolor[HTML]{EFEFEF}0.7 &
  % \cellcolor[HTML]{EFEFEF}0.44 &
  % \cellcolor[HTML]{EFEFEF}0.77 &
  % \cellcolor[HTML]{EFEFEF}0.74 &
  % \cellcolor[HTML]{EFEFEF}0.49 \\ \bottomrule
\end{tabular}

\caption{The accuracies and class-weighted F1 scores of predicting all 143 MARPOR/CMP categories and 3 RILE-specific categories (\textit{left}, \textit{right}, \textit{other}) in the leave-one-country-out (\xcountry) and old-vs.-new (\xtime) settings using a multilingual encoder (\xlmen) or preprocessing via machine translation (\mt).  \textsc{Maj} is the majority-class baseline for each setting.
% The majority-class accuracy baseline for the \xcountry\ setting is 0.59; for the \xtime\ setting, 0.63.
}\label{tab:results-categories}
\end{table}

% \begin{table*}[t]
% \centering
% \small
% \begin{tabular}{lllll}
% \hline
%  & \multicolumn{1}{l}{\begin{tabular}[c]{@{}l@{}}\xcountry\ \\ \xlmen \end{tabular}} & \multicolumn{1}{l}{\begin{tabular}[c]{@{}l@{}}\xcountry\ \\ \mt \end{tabular}} & \multicolumn{1}{l}{\begin{tabular}[c]{@{}l@{}}\xtime\ \\ \xlmen \end{tabular}} & \multicolumn{1}{l}{\begin{tabular}[c]{@{}l@{}}\xtime\ \\ \mt \end{tabular}} \\ \cline{2-5} 
% RILE (\textsc{marpor}) & \cellcolor[HTML]{EFEFEF}\textbf{0.73} & \cellcolor[HTML]{EFEFEF}0.71 & 0.88 & 0.84  \\
% RILE (\textsc{3-way}) & \cellcolor[HTML]{EFEFEF}0.72 & \cellcolor[HTML]{EFEFEF}0.72 & \textbf{0.90} & 0.88 \\
% GAL–TAN (\textsc{marpor}) & \cellcolor[HTML]{EFEFEF}0.78 & \cellcolor[HTML]{EFEFEF}0.79 & 0.81  & 0.83 \\
% GAL–TAN (\textsc{3-way}) & \cellcolor[HTML]{EFEFEF}\textbf{0.80} & \cellcolor[HTML]{EFEFEF}0.78 & 0.80  & \textbf{0.84} \\ \hline
% \end{tabular}
% \caption{The results (Spearman correlations coefficients) of computing RILE and \GT\ scores via 
% predicting all MARPOR categories (rows 1 and 3) vs.\ via predicting metric-specific aggregated in the three-way categories (rows 2 and~4). The best results per scale and setting are in bold. }\label{tab:results-correlations}
% \end{table*}

\begin{table}[t]
  \setlength{\tabcolsep}{3pt}
\small
\centering
\begin{tabular}{@{}llcc@{}}
\toprule
 &  & \multicolumn{1}{l}{RILE (CMP)} & \multicolumn{1}{l}{RILE (3-way)} \\ \cmidrule(l){3-4} 
\multirow{4}{*}{\textsc{x-country}} & \textsc{xlm} & \textbf{0.73} & 0.72 \\
 & \textsc{mt} & 0.71 & 0.72 \\
 & \textsc{bb} & \multicolumn{2}{c}{0.55} \\
 & \textsc{lf} & \multicolumn{2}{c}{0.16} \\ \cmidrule(l){3-4}
\multirow{4}{*}{\textsc{x-time}} & \textsc{xlm} & 0.88 & \textbf{0.9} \\
 & \textsc{mt} & 0.84 & 0.88 \\
 & \textsc{bb} & \multicolumn{2}{c}{0.71} \\
 & \textsc{lf} & \multicolumn{2}{c}{0.35} \\ \bottomrule
\end{tabular}
\caption{The results (Spearman correlations) of computing RILE via 
predicting all MARPOR/CMP sentence-level categories (CMP), RILE-specific categories (3-way), or using LITs (BB: BigBird; LF: Longformer).}\label{tab:results-correlations}
\end{table}

\subsection{Predicting MARPOR categories}\label{ssec:results-categories}

As Table~\ref{tab:results-categories} shows, predicting the
fine-grained MARPOR categories directly is a very hard task, both in
the \xcountry\ and \xtime\ settings. Our models easily beat the
majority-class baseline but only achieve an accuracy above 50\% in the
\xtime\ setting with the \xlmen\ encoder.

Aggregating labels into the three RILE-relevant classes makes the task
predictably simpler: the baseline F1 score rises from nearly zero to
0.44/0.49 (Other becomes the dominant category), but so does the
performance of the models, to accuracies and F1 scores of 0.7
and above. However, there is still ample room for
improvement. Interestingly, while using machine translation leads to
consistent improvements in the \xcountry\ setting, the \xtime\ setting
is better served with the multilingual encoder.\footnote{The results
of using XLM-RoBERTa base in the \xtime\ setting are as follows: 
CMP labels: accuracy~-- 0.51, F1~-- 0.51, \textit{r}~-- 0.87; 3 labels: 
accuracy~-- 0.76, F1~-- 0.75, \textit{r}~-- 0.88.}

\subsection{Computing RILE scores}\label{ssec:results-scales}

% \begin{figure}[t]
%     \centering
%     \includegraphics[width=0.4\textwidth]{img/rile_long_input_barplots_vertical-crop.pdf}
%     \caption{
%     By-country Spearman correlations of gold RILE scores and those predicted using \mt\ with label aggregation
%     (light grey), BigBird (dark grey), and Longformer (black; negative correlations capped at zero).} \label{fig:results-correlations-by-country}
% \end{figure}

\paragraph{Label aggregation}
In agreement with our working hypothesis,
Table~\ref{tab:results-correlations} shows that even noisy labels can
be used to calculate manifesto-wide scale values that are largely in
agreement with gold values. When predicting RILE via label
aggregation the best results are attained by using the
multilingual encoder, both in the \xcountry\ and in the \xtime\
setting. 

Somewhat surprisingly, aggregating the labels, even though
this leads to a small number of surface-level classification mistakes,
does not improve the eventual RILE scores in the \xcountry\ setting
(\textit{r}~= 0.72 from aggregated labels vs.\ 0.73 from all labels)
and gives only a modest boost in the \xtime\ setting (0.9 vs.\ 0.88).

\paragraph{Long-input Transformers}

The performance of LITs is vastly uneven. In the \xcountry\ setting,
both models struggle: by-chunk RILEs from Longformer are essentially
uncorrelated with gold ones, while BigBird's predictions show a
non-negligible correlation (0.55), which is still much worse than the
label aggregation results. In the \xtime\ setting, while Longformer's
predictions are still extremely noisy (\(r = 0.35\)), BigBird's ones
are comparable to what the label aggregation approach achieves in the
\xcountry\ setting (0.71). As we discuss below, however, this
correlation is somewhat misleading: while producing scores that are
monotonically aligned with correct ones, BigBird predicts values that
are very close to zero and thus differ greatly in their absolute
values from the gold scores.

\paragraph{LIT-based classifiers}

\begin{table}[t]
\small
\centering
\begin{tabular}{@{}llcc@{}}
\toprule
 &  & \textsc{x-country} & \textsc{x-time} \\ \cmidrule(l){3-4} 
\multirow{2}{*}{BigBird} & Acc & 0.69 / 0.73 & 0.74 / 0.71  \\
 & F1 & 0.68 / 0.71 & 0.72 / 0.68 \\
\multirow{2}{*}{Longformer} & Acc & 0.59 / 0.66 & 0.58 / 0.64  \\
 & F1 & 0.56 / 0.63 & 0.53 / 0.59 \\ \bottomrule
\end{tabular}
\caption{Performance (on the chunk/manifesto level) of long-input Transformers on the task of 5-way 
political-stance classification. F1 scores are macro averaged and weighted by the frequency of the gold 
classes.}
\label{tab:lit-classification-results}
\end{table}

\begin{table}[t]
\centering
\small
\begin{tabular}{lrrrrr}
\toprule
{} &  L &  CL &  C &  CR &  R \\
\cmidrule(l){2-6}
L   &     \textbf{0} &            3 &         2 &             0 &      0 \\
CL  &     0 &          \textbf{133} &       135 &             1 &      0 \\
C   &     0 &           69 &       \textbf{708} &            41 &      0 \\
CR  &     0 &            0 &        70 &            \textbf{28} &      0 \\
R   &     0 &            0 &         1 &             1 &      \textbf{0} \\
\bottomrule
\end{tabular}
\caption{Confusion matrix for the party stance predicted by the BigBird-based
classifier in the \xcountry\ setting. L: left, CL: centre left, C: centrist, 
CR: center right, R: right.}
\label{tab:bigbird-classifier-cm}
\end{table}

The results of the application of the better-performing LIT, BigBird,
to the task of 5-way stance classification are shown in
Table~\ref{tab:lit-classification-results}. Unlike RILE scores,
by-chunk stance labels cannot be averaged, so for the final prediction
each manifesto is assigned its majority class. The performance of the
BigBird-based model in this setting is reasonable, with F1 scores
\(\approx 0.7\).

\subsection{Error analysis}\label{ssec:results-errors}

\subsubsection{Regression to the mean}

\begin{figure*}[t]
    \centering
    \includegraphics[width=\textwidth]{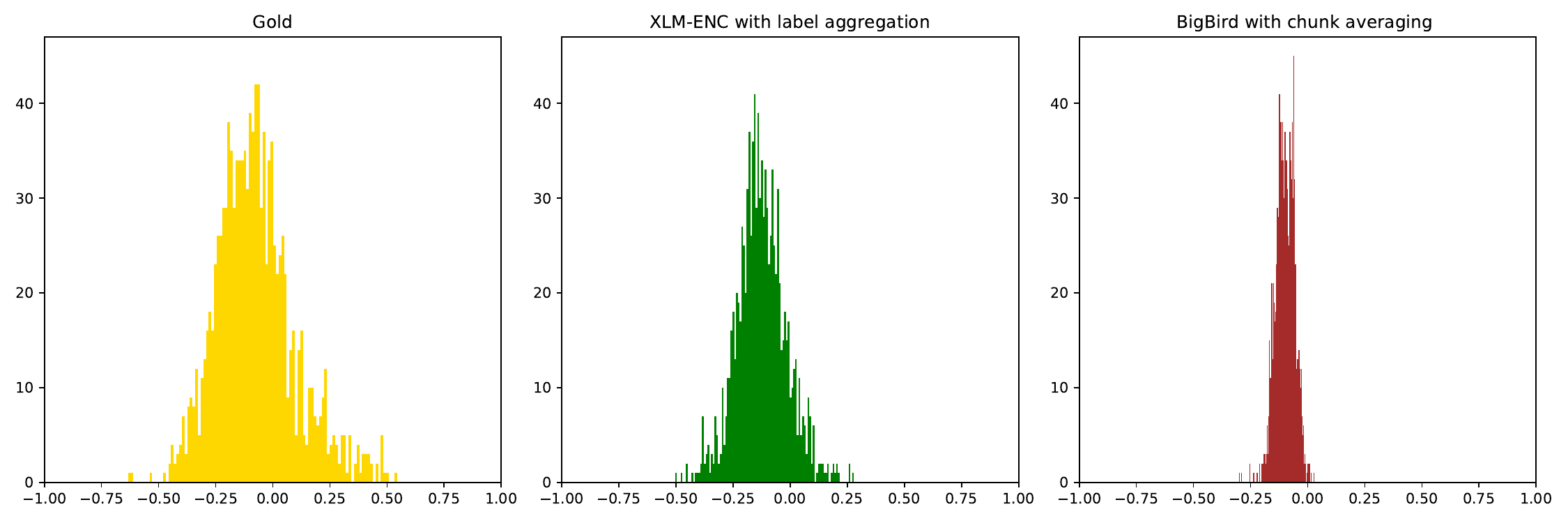}
    \caption{The distributions of gold and predicted RILE scores in the \xcountry\ setting.}
    \label{fig:me_rile_histograms}
\end{figure*}

The distributions of gold RILE scores and those predicted in the \xcountry\ setting by the
best-performing label-aggregation pipeline and the best-performing LIT are shown in 
Figure~\ref{fig:me_rile_histograms}.\footnote{The situation in the \xtime\ setting is similar.
The corresponding plots are presented in Appendix~\ref{app:ovn-rile-histograms}.}
The plots make it clear that both models are very conservative:
predicted values cluster closer to the mean RILE score than in the gold data. BigBird is especially
affected by this, which we take to indicate that it suffers from a lack of training data:
the training dataset was big enough to correctly estimate the mean of the distribution but not
big enough to approximate the correct dispersion.

The predictions of the label-aggregation model based on \xlmen\ approximate the dispersion much
better. However, the model still fails to account for the heavy right tail in the gold data
and presents a more symmetric picture. In terms of RILE scores, this corresponds to a \textit{left 
skew}: the model often presents right-leaning manifestos (those with positive RILE scores) as more centrist.

\begin{figure}[th]
    \centering
    \includegraphics[width=0.49\textwidth]{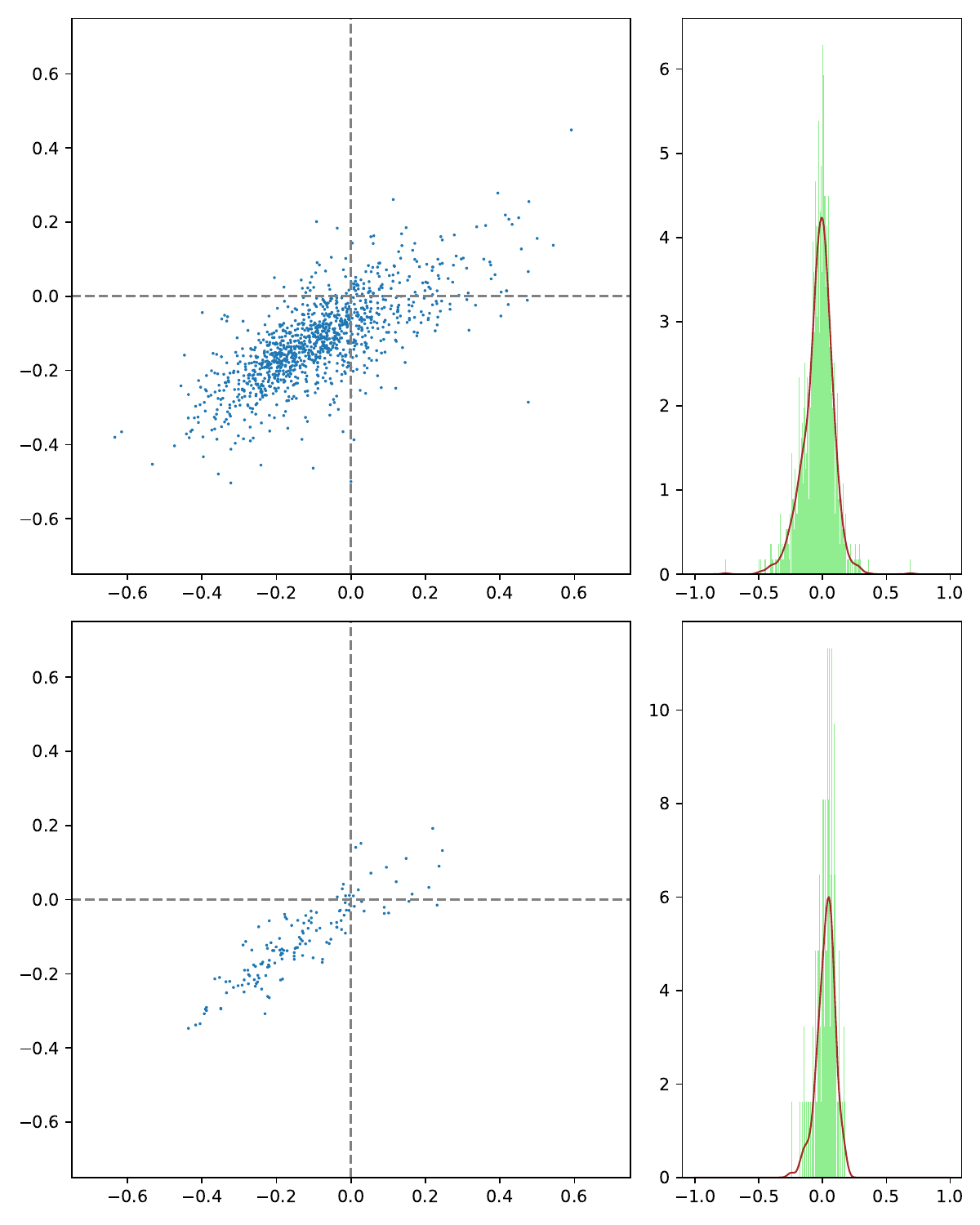}
    \caption{Gold vs.\ predicted RILE scores and histograms with density contours of the prediction 
    errors in the \xcountry\ (top) and \xtime\ (bottom) settings with \xlmen\ and label aggregation.}
    \label{fig:me_rile_errors}
\end{figure}

A~more detailed picture of the relationship between the gold RILE scores and those predicted 
by the label-aggregation model is shown, for both settings, in Figure~\ref{fig:me_rile_errors}, 
which also presents the density of the prediction errors. Consistently with 
Figure~\ref{fig:me_rile_histograms}, the density of the \xcountry\ error distribution has a slightly 
heavier left tail. To characterize this behavior, we can look at the cases where the sign of the prediction is flipped, 
i.e.\ the upper-left and the lower-right quadrants of the scatterplot. 
While the UL quadrant is nearly empty, the LR quadrant is populated not only near the $x = 0$  
asymptote, but also further to the right. This suggests that
in the cross-country and cross-lingual setting, the hardest aspect of the problem  
is correct identification of right-wing statements across countries. 

One of the challenges associated with right-wing labels
are their differing distributions across countries.
While the variation in the cumulative share of left-wing labels in manifestos
% (averaged across all manifestos)
is bounded roughly between 0.2 and 0.3, 
with the same labels dominant everywhere, the variability of 
right-wing labels is much higher and their share is lower on average.
See Figure~\ref{fig:right_left_across_countries} in 
Appendix~\ref{app:sec:cum-shares} for details and 
\citet{lachat2018left,fielitz2021right,jahn2022left} for more in-depth analyses.

As the bottom panel of Figure~\ref{fig:me_rile_errors} shows, the
magnitude of errors in the \xtime\ setting is considerably lower, with
only a handful of sign-flip errors. This indicates that when a model
has access to in-country data, the estimation of political
positioning becomes easier, and the identification of
right-wing tendencies is not a major hurdle any more.

The 5-way LIT-based party-stance classifier also suffers from the
regression-to-the-mean problem, as can be seen in
Table~\ref{tab:bigbird-classifier-cm}: the centrist category is
overpredicted, while two extreme categories, which are rare
in the data, are never predicted correctly.

% As for the GAL-TAN scale, the difficulty of identifying authoritarian manifestos persists, while libertarian ones have hardly any errors (Fig.~\ref{fig:mt_ovn_results_gt}). 

% We return to this 
% question and provide a more extensive by-country overview in the Discussion section 
% below.

\subsubsection{Classifier errors and scaling analysis}

One of the surprising results in Tables \ref{tab:results-categories} and \ref{tab:results-correlations}
is that low accuracy of the models trying to predict all MARPOR labels directly does not
translate into low quality of respective RILE scores in the \xcountry\ setting. This
seems to suggest that errors of the models are not random: the models rather substitute, e.g.,
another Left-category label for a true Left-category label than replace a label from the Left
set with a label from the Right set. A~confusion matrix for the 3 coarse-grained labels 
(computed based on the \textit{fine-grained} labels predicted by \xlmen\  in the 
\xcountry\ setting) shown in Table~\ref{tab:rile-cm} demonstrates that this is indeed the case. 

\begin{table}[t]
\centering
% \small
\begin{tabular}{lrrr}
\toprule
 & Right & Left & Other \\  \cmidrule(l){2-4} 
Right & \textbf{46} & 20 & 33 \\
Left & 8 & \textbf{66} & 26 \\
Other & 9 & 16 & \textbf{75} \\ \bottomrule
\end{tabular}
\caption{Confusion matrix of coarse-grained labels used to compute the RILE
score based on all MARPOR labels (the \xlmen\ + \xcountry\ setting). True labels are in the
rows, predicted labels in the columns.}\label{tab:rile-cm}
\end{table}

\section{Discussion}\label{sec:discussion}

Our results show that multi-lingual automatic analysis of
political-party positioning is at least partially feasible.
It is possible to provide a high-level overview of the party system in a new
country with a reasonable degree of precision, and even better results
can be achieved with some amount of in-country data: the RILE scores
computed using our method demonstrate a remarkably high correlation
with the gold scores. Interestingly, the main obstacle to the success of
our method seems to be not the language barrier, which is bridged well
by either the off-the-shelf MT systems or the multilingual
encoder, but the differences in the political culture across
countries: the models struggle to correctly identify right-wing
statements in the manifestos.

% One of the reasons why the correct identification of right-wing
% labels is a challenge may be differing distributions of labels across
% countries. While the variation in the cumulative share of left-wing labels across 
% countries 
% % (averaged across all manifestos)
% is bounded roughly between 0.2 and 0.3, 
% with the same labels dominant everywhere, the cumulative share and variability of 
% right-wing labels in the same set of countries is much more varied. See Figure~\ref{fig:right_left_across_countries} in the Appendix for details.

% This does not explain why individual labels may be hard to predict, as they are
% not unique to individual countries but it may be conceivable that countries with a high
% cumulative share of, for example, right-wing labels have a higher thematic variety \textit{inside}
% the categories (such as immigration, traditional way of life, or military) than countries with a 
% more left-wing outlook, focused on a more stable set of social-policy issues \citep{noury2020}.
% Testing this hypothesis is outside of the scope of this study.

% Finally, even though the results of the Spearman correlation in the label-aggregation setting are very 
% similar between the MARPOR and the three-way setups (Table \ref{tab:results-correlations}), the best 
% performance is reached by the latter in 3 out of the 4 settings, which suggests that label aggregation is 
% useful for the downstream use of our approach.

In practical terms, using long-input Transformers instead of
sentence-level classifiers offers a way to greatly simplify the
analysis and obviate the problems of subsentence identification in the
input, as such models are able to make holistic judgements about long
spans of text. In terms of performance, LITs struggle on the task of
directly estimating RILE, compared to label-aggregation models, with
the best model only approaching a reasonable level of
performance. However, this must be taken with a grain of salt, since
the label aggregation models have the advantage of gold-statement
boundaries. Furthermore, our binned-regression experiment shows that
LITs are promising candidates for coarse-grained party positioning
analysis in terms of political \enquote{camps}. For all models, the tails of
the distribution remain hard to identify, with extreme
categories rarely predicted correctly and centre left/centre right labels
often mistaken for centrist.

\section{Related work}\label{sec:related-work}

The work on computational analysis of political documents traditionally
employs bag-of-words methods, such as those popularised by 
\citet{laver2003extracting} and \citet{slapin2008scaling}. \citet{glavas-etal-2017-unsupervised}
introduce distributional semantics in the left--right analysis by using multilingual
word alignment in the embedding space and a graph-based score-propagation algorithm.
This approach is then built upon by \citet{nanni2022political}.

\citet{rheault2020word} adapt the word2vec methodology to the analysis of parliamentary
speeches in a single-language setting via the use of trained party vectors, whose dimensionality they 
reduce using PCA; they then interpret one of the resulting axes as the left--right scale. 
\citet{vafa-etal-2020-text} instead develop a methodology for identifying the political position of 
lawmakers on the progressive-to-moderate dimension with a bag-of-words-based topic-modelling approach. 

The use of contextualised embeddings for political analysis has not yet become mainstream.
\citet{abercrombie-etal-2019-policy} test a wide range of methods, from unigram statistics to
BERT-based classifiers, for assigning MARPOR labels to classify debate motions from the UK parliament.
\citet{dayanik-etal-2022-improving} use several pre-trained single-language BERT models for the task of 
political-statement classification in five languages. Facing the same issues of label-frequency imbalance 
and rare labels, they mitigate them to some degree by using the hierarchical organisation of MARPOR 
labels; they do not try to compute RILE scores.

\citet{ceron-2022} introduce sentence transformers \citep{reimers-gurevych-2019-sentence}
into the problem space and fine-tune the embedding model itself in order to learn a politically 
informative distance measure between manifesto texts. \citet{ceron-etal-2023-additive} 
further extend this method to analyse inter-party differences with regard to major policy 
domains, such as Law and Order or Sustainability and Agriculture.

More generally, our work falls into the domain of zero-shot classification with test data 
coming from a country or a time period not covered by the training data. The question of
whether machine translation \citep{schafer-etal-2022-cross} or multilingual encoders 
\citep{litschko2022cross} is better suited for cross-lingual transfer is still actively debated,
and we explore both options. From another perspective, the task of identifying and 
characterising political positions from textual data abuts larger fields of 
stance detection and argument mining \citep{kuccuk2020stance, reimers-etal-2019-classification}.

\section{Conclusion}\label{sec:conclusion}

In this paper, we have proposed the first series models that
generalise the task of
%showed that is possible to conduct fully automatic analysis of 
political-party positioning across countries and election cycles. 
% Previous work in this area was 
% mostly limited to bag-of-words-based approaches or small samples,
% \citep{laver2003extracting,slapin2008scaling,vafa-etal-2020-text}, 
% instead of directly predicting MARPOR category labels in the whole set of target manifestos in order to 
% perform the scaling analysis, as it is done in the traditional political-science
% literature. 
We showed that the main challenge -- predicting MARPOR
labels across countries and election cycles with high accuracy -- is, surprisingly, not a real
barrier on the way to a highly precise multi-lingual scaling analysis. We experimented with
the Standard Right--Left Scale (RILE score), which is widely discussed in the
political-science literature, and demonstrated that party manifestos can be effectively characterized
in these terms using state-of-the-art multilingual modeling techniques applied to sentence-level
classification with subsequent label aggregation and that even better results can be achieved via 
task-specific label clustering.

We further experimented with replacing the label-aggregation approach
with long-input Transformers -- both using regression and
classification formulations -- in order to obviate the task of
identifying spans of statements from manifestos. These models
demonstrate promising performance but still underperform
the more traditional pipeline mimicking manual analysis.

Bridging the gap between long-input models and political analysis is an
important avenue for future work, together with tackling other political
dimensions and further widening the scope of the analysis.

\FloatBarrier

% We extend previous work in two ways. On one hand, we conduct the first analysis of the whole
% annotated MARPOR dataset showing that it is possible to conduct political analysis at scale:
% previous approaches targeted a single country or a handful of countries. On the other hand,
% we directly employ state-of-the-art sentence-embedding models in order to produce 
% MARPOR-label-based scores for manifestos. This generalises previous analyses and
% gives a more interpretable scaling procedure.

\section*{Limitations}

% We must note that our method tends to somewhat underestimate right-wing and authoritarian tendencies in
% manifestos, which is most likely due to higher thematic variation inside right-wing issues as
% defined by the MARPOR categories (Table \ref{tab:rile-categories}). A~more effective analysis of right-wing 
% statements is an important avenue for future work. 

% Another angle that we have not explored in this study is
% \enquote{backwards} analysis: the identification of policy categories in early texts.
% The MARPOR annotations only become sufficiently ample in the 21st centuries. The category
% set itself, however, can be argued to be applicable starting from 1930s. The degree to which
% state-of-the-art embedding models can tackle earlier political texts given the semantic
% drift in the corpora \citep{dubossarsky-etal-2017-outta} is of much interest to the political-science community.

The main limitations of our work are twofold, and both stem from our dependence on the categories
and annotations produced by the MARPOR project:

\begin{enumerate}
    \item The RILE scale that we target is computed based on the MARPOR
    category labels, and we do not test if our methodology can be easily projected to other
    categorisation schemes. However, given the important role of the MARPOR codebook in the
    political-science literature and the amount of annotated data already available, we hope 
    that our work makes a valuable contribution to the debate.
    \item In label-aggregation pipeline, we are dependent not only on the labels themselves but also on 
    the way they are applied to manifestos: following previous work 
    \citep{dayanik-etal-2022-improving,ceron-2022}, we use the sub-sentence boundaries selected
    by MARPOR annotators in order to assign a single category to each statement. In the 
    manifesto texts, sentences therefore sometimes can be associated with several labels. There
    are several possible ways to address this issue (e.g., selecting a \enquote{majority} label for
    each sentence in the training data, training a multi-label classifier, or learning splits together
    with labels from the training set), and they need to be explored to obtain best possible
    performance in real-world settings. Using LITs removes this issue, but their performance is
    not competitive.
\end{enumerate}

%\section*{Limitations}
% required by ACL policy

% \section*{Ethics Statement}
% required by ACL policy

\section*{Acknowledgments}

We acknowledge partial support by Deutsche Forschungsgemeinschaft (DFG) for project MARDY~2 (375875969) within priority program RATIO.

\bibliography{anthology,custom}
\bibliographystyle{acl_natbib}

%\onecolumn
\appendix

\section{Examples of the MARPOR categories}
\label{sec:appendix-a}

See Table~\ref{tab:manifestos}.

\begin{table*}[th!]
\centering
\begin{tabular}{lp{10cm}p{3cm}}
\hline
\textbf{Party} & \textbf{Text}                                                                                                                                                 & \textbf{Category}                                                        \\ \hline
AfD            & The principles of equality before the law.                                                                                                                    & Equality: Positive                                                       \\
CDU            & We are explicitly committed to NATO's  2\% target.                                                                                                            & Military: Positive                                                       \\
  FDP            & And with a state that is strong because it acts lean and modern instead of complacent, old-fashioned and sluggish. &
                                                                                                                                        Governm. and Admin. Efficiency \\
SPD            & There need to be alternatives to the big platforms - with real opportunities for local suppliers.                  & Market Regulation                                                        \\
Grüne          & We will ensure that storage and shipments are strictly monitored.                                                                                             & Law and Order: Positive                                                  \\
Die Linke      & Blocking periods and sanctions are abolished without exception.                                                                                               & Labour groups: Positive                                                  \\ \hline
\end{tabular}
\caption{Translated examples of sentences from German federal election manifestos (2021)  with their categories as annotated by the Comparative Manifesto Project.}
\label{tab:manifestos}
\end{table*}

\section{Names of MARPOR categories referenced by a number in the text}
\label{app:marpor-categories}

\begin{itemize}
    \setlength{\itemindent}{2em}
    \item[103] Anti-Imperialism
    \item[104] Military: Positive
    \item[105] Military: Negative
    \item[106] Peace
    \item[107] Internationalism: Positive
    \item[201] Freedom and Human Rights
    \item[201.1] Freedom
    \item[201.2] Human Rights
    \item[202] Democracy General
    \item[202.1] Democracy General: Positive
    \item[203] Constitutionalism: Positive
    \item[305] Political Authority
    \item[401] Free Market Economy
    \item[402] Incentives: Positive
    \item[403] Market Regulation
    \item[404] Economic Planning
    \item[406] Protectionism: Positive
    \item[406.1] Anti-Growth Economy:Positive
    \item[407] Protectionism: Negative
    \item[412] Controlled Economy
    \item[413] Nationalisation
    \item[414] Economic Orthodoxy
    \item[416] Anti-Growth Economy: Positive
    \item[501] Environmental Protection
    \item[502] Culture: Positive
    \item[504] Welfare State Expansion
    \item[505] Welfare State Limitation
    \item[506] Education Expansion
    \item[601] National Way of Life: Positive
    \item[602] National Way of Life: Negative
    \item[603] Traditional Morality: Positive
    \item[604] Traditional Morality: Negative
    \item[605] Law and Order
    \item[605.1] Law and Order: Positive
    \item[606] Civic Mindedness: Positive
    \item[607] Multiculturalism: Positive
    \item[608] Multiculturalism: Negative
    \item[701] Labour Groups: Positive
    \item[705] Unprivileged Minority Groups
    \item[706] Non-economic Demographic Groups
\end{itemize}

\section{Dataset breakdown by country and by language}
\label{app:dataset-stats}

See Tables \ref{tab:dataset-stats} and \ref{tab:dataset-stats-languages}.

% \usepackage{booktabs}
\begin{table*}[t]
\centering
\begin{tabular}{@{}lll@{}}
\toprule
Country & \# manifestos & \# sentences \\ \midrule
Argentina & 29 & 10983 \\
Armenia & 22 & 1623 \\
Australia & 30 & 21683 \\
Austria & 32 & 39452 \\
Belgium & 43 & 154699 \\
Bolivia & 9 & 7718 \\
Bulgaria & 18 & 8945 \\
Canada & 23 & 28524 \\
Chile & 17 & 33988 \\
Czech Republic & 31 & 25986 \\
Denmark & 45 & 17073 \\
Estonia & 23 & 16524 \\
Finland & 33 & 22520 \\
France & 20 & 9347 \\
Georgia & 19 & 2610 \\
Germany & 35 & 81759 \\
Hungary & 26 & 45246 \\
Iceland & 34 & 8139 \\
Ireland & 23 & 30348 \\
Israel & 1 & 24 \\
Italy & 32 & 22091 \\
Japan & 9 & 3387 \\
Latvia & 30 & 2030 \\
Luxembourg & 17 & 30768 \\
Mexico & 48 & 49818 \\
Netherlands & 45 & 72610 \\
New Zealand & 44 & 43869 \\
North Macedonia & 43 & 56719 \\
Poland & 30 & 27285 \\
Russia & 4 & 1350 \\
Slovakia & 33 & 25325 \\
South Africa & 24 & 12835 \\
South Korea & 5 & 6030 \\
Spain & 90 & 142878 \\
Sweden & 31 & 17293 \\
Switzerland & 50 & 20975 \\
Turkey & 23 & 54472 \\
Ukraine & 35 & 3099 \\
United Kingdom & 32 & 33211 \\
United States & 9 & 16262 \\
Uruguay & 5 & 16518 \\ \bottomrule
\end{tabular}
\caption{Number of manifestos and number of sentences per country.}\label{tab:dataset-stats}
\end{table*}

% \usepackage{booktabs}
\begin{table*}[t]
\centering
\begin{tabular}{@{}lllll@{}}
\toprule
Language code & Language & \# manifestos & \# sentences & Countries \\ \midrule
bg & Bulgarian & 18 & 8945 & Bulgaria \\
ca & Catalan & 18 & 32780 & Spain \\
cs & Czech & 31 & 25986 & Czech Republic \\
da & Danish & 45 & 17073 & Denmark \\
de & German & 123 & 163622 & \begin{tabular}[c]{@{}l@{}}Austria, Germany, Italy, \\ Luxembourg, Switzerland\end{tabular} \\
en & English & 177 & 171812 & \begin{tabular}[c]{@{}l@{}}Australia, Canada, Ireland, Israel, \\ New Zealand, South Africa, \\ United Kingdom, United States\end{tabular} \\
es & Spanish & 174 & 223047 & \begin{tabular}[c]{@{}l@{}}Argentina, Bolivia, Chile, \\ Mexico, Spain, Uruguay\end{tabular} \\
et & Estonian & 23 & 16524 & Estonia \\
fi & Finnish & 29 & 21313 & Finland \\
fr & French & 52 & 105570 & \begin{tabular}[c]{@{}l@{}}Belgium, Canada, France, \\ Luxembourg, Switzerland\end{tabular} \\
gl & Galician & 6 & 6076 & Spain \\
hu & Hungarian & 26 & 45246 & Hungary \\
hy & Armenian & 22 & 1623 & Armenia \\
is & Icelandic & 34 & 8139 & Iceland \\
it & Italian & 33 & 21646 & Italy, Switzerland \\
ja & Japanese & 9 & 3387 & Japan \\
ka & Georgian & 19 & 2610 & Georgia \\
ko & Korean & 5 & 6030 & South Korea \\
lv & Latvian & 30 & 2030 & Latvia \\
mk & Macedonian & 43 & 56719 & North Macedonia \\
nl & Dutch & 75 & 155807 & Belgium, Netherlands \\
pl & Polish & 30 & 27285 & Poland \\
ru & Russian & 4 & 1350 & Russia \\
sk & Slovak & 33 & 25325 & Slovakia \\
sv & Swedish & 35 & 18500 & Finland, Sweden \\
tr & Turkish & 23 & 54472 & Turkey \\
uk & Ukrainian & 35 & 3099 & Ukraine \\ \bottomrule
\end{tabular}
\caption{Number of manifestos and sentences per language and respective source countries.}\label{tab:dataset-stats-languages}
\end{table*}

% \section{Error plots for the \GT\ score}
% \label{error_plots:gt}

% See Figures \ref{fig:me_loco_results_gt} and \ref{fig:mt_ovn_results_gt}.

% \begin{figure*}[t]
%     \centering
%     \includegraphics[width=\textwidth]{img/me_loco_gt.pdf}
%     \caption{A scatterplot of predicted and gold \GT\ scores and a density plot of absolute
%     errors for the \xlmen\ + \xcountry\ setting. The more negative, the more libertarian (GAL) a party is. }
%     \label{fig:me_loco_results_gt}
% \end{figure*}

% \begin{figure*}[t]
%     \centering
%     \includegraphics[width=\textwidth]{img/mt_mpnet_ovn_gt.pdf}
%     \caption{A scatterplot of predicted and gold RILE scores and a density plot of absolute errors for the \mt\ + \xtime\ setting. The more negative, the more libertarian (GAL) a party is. }
%     \label{fig:mt_ovn_results_gt}
% \end{figure*}

\section{Distributions of predicted RILEs in the \xtime\ setting}
\label{app:ovn-rile-histograms}

\begin{figure*}[t]
    \centering
    \includegraphics[width=\textwidth]{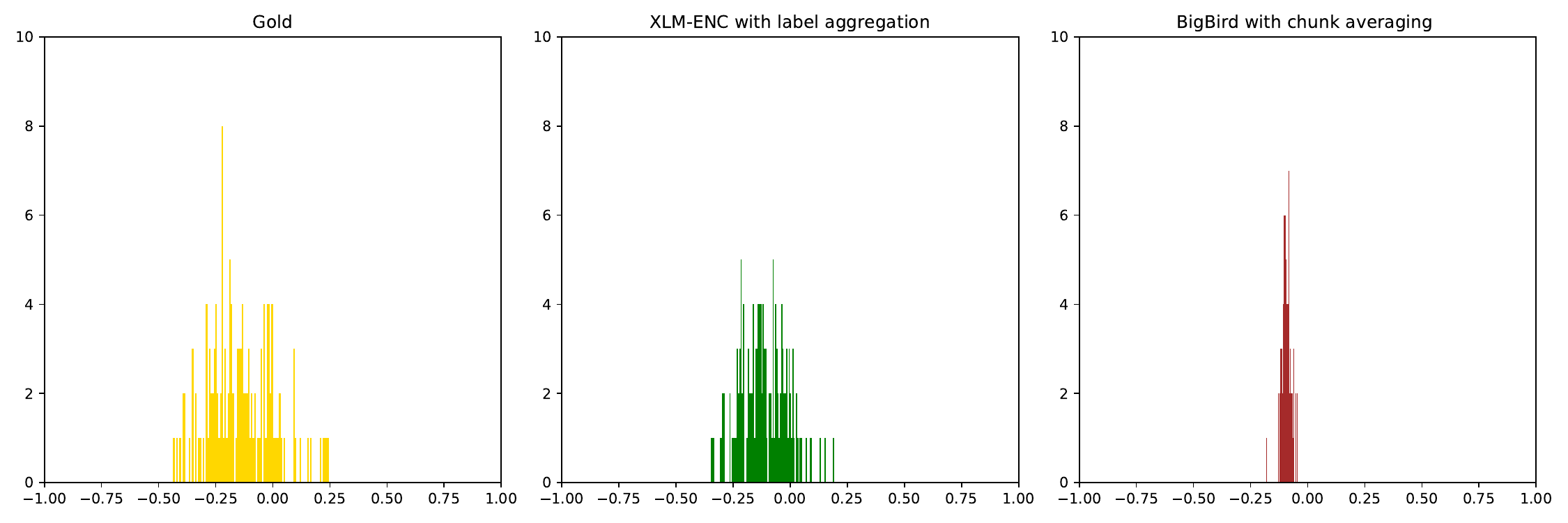}
    \caption{The distributions of gold and predicted RILE scores in the \xtime\ setting.}
    \label{fig:ovn_rile_histograms}
\end{figure*}

See Figure \ref{fig:ovn_rile_histograms}.

\section{Cumulative share of left and right categories across countries}
\label{app:sec:cum-shares}

\begin{figure*}[t]
    \centering
    \includegraphics[width=\textwidth]{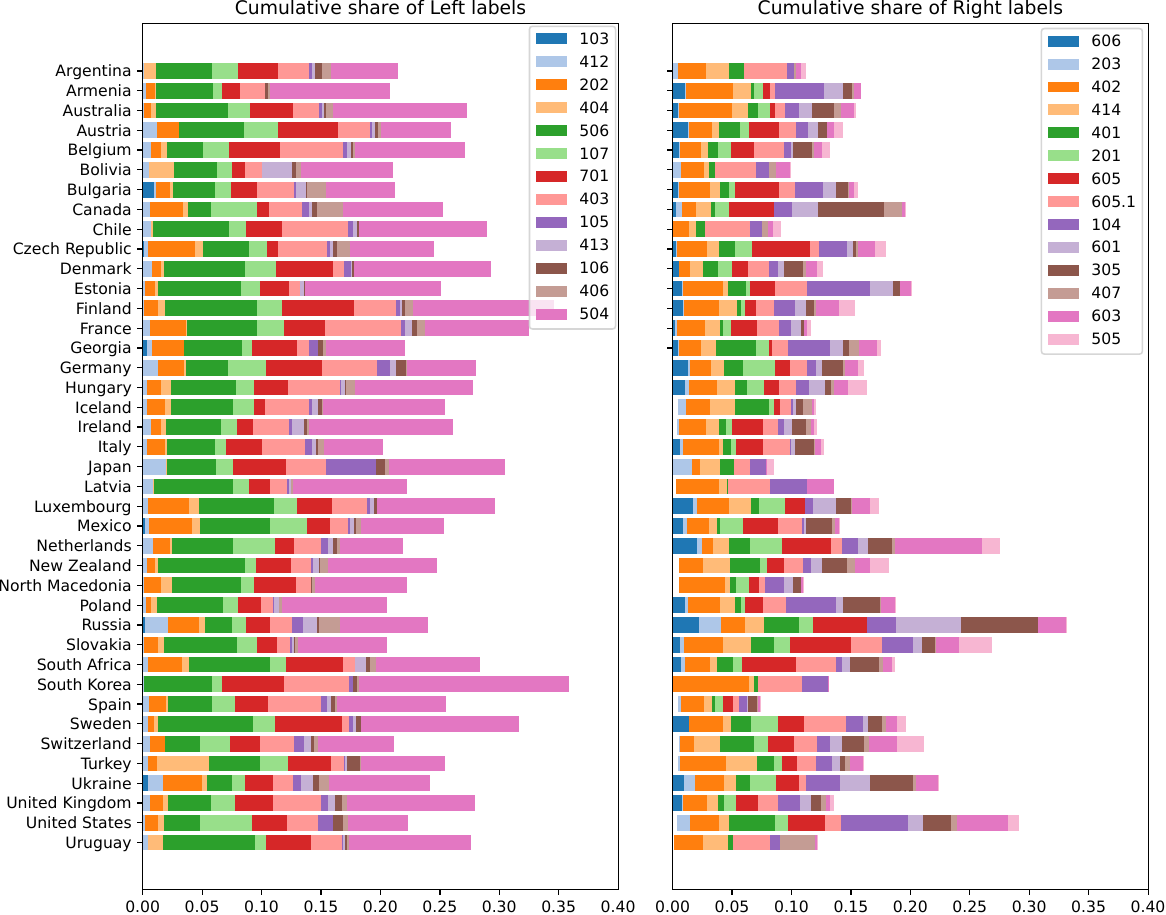}
    \caption{Cumulative shares of left-wing and right-wing labels in manifestos from different countries. See Appendix~\ref{app:marpor-categories} for the explanation of label codes.}
    \label{fig:right_left_across_countries}
\end{figure*}

See Figure \ref{fig:right_left_across_countries}. 

% \begin{figure*}[t]
%     \centering
%     \includegraphics[width=\textwidth]{img/libertarian_authoritarian_labels.png}
%     \caption{Cumulative shares of libertarian and authoritarian labels in manifestos from different countries. See Appendix~B for the explanation of label codes.}
%     \label{fig:cummulative_galtan}
% \end{figure*}

\end{document}